\documentclass[10pt,twocolumn,letterpaper]{article}

\usepackage{wacv}
\usepackage{times}
\usepackage{epsfig}
\usepackage{graphicx}
\usepackage{amsmath}
\usepackage{amssymb}
\usepackage{booktabs}
\usepackage{array}
\usepackage{caption}
\usepackage{multirow}
\usepackage{stmaryrd} %
\usepackage{xcolor}
\usepackage{enumitem}
\usepackage{pdfpages}
\setlist{topsep=0pt,parsep=0pt,itemsep=1pt} %
\graphicspath{{figures/}}
 
\def\wacvPaperID{1367} %

\wacvapplicationstrack %

\wacvfinalcopy %

\ifwacvfinal
\usepackage[breaklinks=true,bookmarks=false]{hyperref}
\else
\usepackage[pagebackref=true,breaklinks=true,colorlinks,bookmarks=false]{hyperref}
\fi

\usepackage[capitalise,nameinlink]{cleveref}
\newcolumntype{C}[1]{>{\centering\arraybackslash}p{#1}}
\newcolumntype{L}[1]{>{\raggedright\arraybackslash}p{#1}}
\newcolumntype{R}[1]{>{\raggedleft\arraybackslash}p{#1}}

\newcommand{\ignore}[1]{}

\newcommand{\defeq}{\mathrel{\mathop:}=}

\newcommand{\std}[1]{\tiny $\pm$#1}

\newcommand{\vlike}{p_\theta \left( \mathcal{X} \,|\, Z, \mathcal{C}, T \right)}
\newcommand{\vpost}{q_\phi \left( Z \,|\, \mathcal{X}, \mathcal{C}, T \right)}
\newcommand{\vpri}{p \left( Z \,|\, \mathcal{C}, T \right)}
\newcommand{\vmu}{\mu \left( \mathcal{X}, \mathcal{C}, T \right)}
\newcommand{\vsigma}{\sigma \left( \mathcal{X}, \mathcal{C}, T \right)}

\begin{document}

\title{Generative Colorization of Structured Mobile Web Pages}

\author{
  Kotaro Kikuchi$^1$
  \and
  Naoto Inoue$^1$
  \and
  Mayu Otani$^1$\\[.5em]
  \hspace{-8em}$^1$CyberAgent
  \and
  Edgar Simo-Serra$^2$\\[.5em]
  \hspace{-2em}$^2$Waseda University
  \and
  Kota Yamaguchi$^1$
}

\maketitle
\thispagestyle{empty}

\begin{abstract}
   Color is a critical design factor for web pages, affecting important factors
such as viewer emotions and the overall trust and satisfaction of a website.
Effective coloring requires design knowledge and expertise, but if this process
could be automated through data-driven modeling, efficient exploration and
alternative workflows would be possible.  However, this direction remains
underexplored due to the lack of a formalization of the web page colorization
problem, datasets, and evaluation protocols.  In this work, we propose a new
dataset consisting of e-commerce mobile web pages in a tractable format, which
are created by simplifying the pages and extracting canonical color styles with
a common web browser.  The web page colorization problem is then formalized as a
task of estimating plausible color styles for a given web page content with a
given hierarchical structure of the elements.  We present several
Transformer-based methods that are adapted to this task by prepending structural
message passing to capture hierarchical relationships between elements.
Experimental results, including a quantitative evaluation designed for this
task, demonstrate the advantages of our methods over statistical and image
colorization methods.  The code is available at
\url{https://github.com/CyberAgentAILab/webcolor}.

\end{abstract}

\newcommand{\MainTable}{
\begin{table*} \begin{center}

\caption{Quantitative comparison of color style generation. The methods with
dagger symbols ($\dagger$) generate in a deterministic manner, while the others
generate in a stochastic manner. The Fr\'echet Color Distances
are multiplied by 1e-3 for clarity.
Note that only ColTran uses the ground-truth grayscale screenshot, \ie, partial
color information per pixel.}
\label{tab:main_result}

{\footnotesize \begin{tabular}{l cc cc ccc cc} \toprule
& \multicolumn{2}{c}{Accuracy}
& \multicolumn{2}{c}{Macro F-score}
& \multicolumn{3}{c}{Fr\'echet Color Distance}
& \multicolumn{2}{c}{Contrast violation}\\

Method
& RGB $\uparrow$ & Alpha $\uparrow$ 
& RGB $\uparrow$ & Alpha $\uparrow$ 
& BG $\downarrow$ & Text $\downarrow$ & Pixel $\downarrow$
& \% Pages & \# Elements \\
\midrule

ColTran~\cite{Kumar21Colorization}
& .285\std{.000} & .411\std{.000}
& .009\std{.001} & .101\std{.000}
& 665.78\std{2.49} & 103.50\std{1.61} & \textbf{3.14}\std{0.58}
& 95.51\std{0.15} & 5.29\std{0.02} \\

Stats (mode)$^\dagger$
& .717 & \underline{.891}
& .003 & .219
& 20.94 & 263.11 & 7.43
& 34.17 & 1.18 \\

Stats (sampling)
& .621\std{.000} & .821\std{.000}
& .004\std{.000} & .207\std{.001}
& \textbf{0.71}\std{0.02} & 82.22\std{0.31} & 169.80\std{1.05}
& 94.41\std{0.08} & 4.47\std{0.02} \\

AR (greedy)$^\dagger$
& .720\std{.002} & .886\std{.002}
& .033\std{.002} & .405\std{.003}
& 2.93\std{0.33} & 39.19\std{2.61} & 22.98\std{3.21}
& 66.43\std{0.70} & 2.02\std{0.02} \\

AR (top-p, p=0.8)
& .717\std{.003} & .885\std{.002}
& .032\std{.002} & .403\std{.002}
& 2.63\std{0.37} & 33.79\std{3.59} & 33.67\std{0.55}
& 71.00\std{1.12} & 2.30\std{0.06} \\

AR (top-p, p=0.9)
& .714\std{.003} & .883\std{.003}
& .030\std{.002} & .402\std{.006}
& 2.40\std{0.33} & 33.00\std{3.81} & 37.44\std{0.30}
& 73.42\std{1.01} & 2.40\std{0.06} \\

NAR$^\dagger$
& \textbf{.773}\std{.001} & \textbf{.929}\std{.001}
& \textbf{.076}\std{.001} & \textbf{.670}\std{.002}
& 1.57\std{0.33} & \textbf{21.81}\std{2.20} & 5.98\std{1.23}
& 74.02\std{0.92} & 2.25\std{0.09} \\

CVAE
& \underline{.771}\std{.001} & \textbf{.929}\std{.000}
& \underline{.069}\std{.001} & \underline{.665}\std{.003}
& \underline{1.50}\std{0.04} & \underline{28.14}\std{1.13} & \underline{4.20}\std{0.19}
& 74.71\std{0.19} & 2.23\std{0.05} \\

\midrule

{\scriptsize CVAE w/o message passing}
& .762\std{.001} & .918\std{.000}
& .062\std{.002} & .620\std{.005}
& 2.16\std{0.03} & 33.02\std{2.82} & 10.58\std{1.29}
& 75.10\std{0.79} & 2.29\std{0.06} \\

{\scriptsize CVAE w/o residual connection}
& .768\std{.001} & .927\std{.000}
& .064\std{.000} & .647\std{.009}
& 1.41\std{0.25} & 29.94\std{2.58} & 5.40\std{1.17}
& 74.25\std{0.87} & 2.20\std{0.07} \\

\midrule

Real data
& 1.000 & 1.000
& 1.000 & 1.000
& 0.08 & 3.04 & 0.56
& 71.72 & 2.39 \\

\bottomrule \end{tabular} }
\end{center} \end{table*}
\setlength\tabcolsep{6pt}
}

\newcommand{\showFig}[3]{\frame{\includegraphics[width=20mm]{results/#1/#2/#3.png}}}

\newcommand{\MainFigure}{
\setlength{\tabcolsep}{3pt}
\begin{figure*}
\begin{center}
{\small \begin{tabular}{
    C{20mm}C{20mm}C{20mm}C{20mm}
    C{20mm}C{20mm}C{20mm}C{20mm}
    }
    \showFig{gt}{22}{0}
    & \showFig{mode}{22}{0}
    & \multicolumn{2}{c}{\showFig{sampling}{22}{0} \showFig{sampling}{22}{2}}
    & \showFig{gt}{0}{0}
    & \showFig{mode}{0}{0}
    & \multicolumn{2}{c}{\showFig{sampling}{0}{0} \showFig{sampling}{0}{5}}
    \\
    Ground-truth
    & Stats (mode)
    & \multicolumn{2}{c}{Stats (sampling)}
    & Ground-truth
    & Stats (mode)
    & \multicolumn{2}{c}{Stats (sampling)} \\[2mm]

    \showFig{coltran_gray}{22}{0}
    & \showFig{coltran_color}{22}{0}
    & \multicolumn{2}{c}{\showFig{coltran}{22}{0} \showFig{coltran}{22}{1}}
    & \showFig{coltran_gray}{0}{0}
    & \showFig{coltran_color}{0}{15}
    & \multicolumn{2}{c}{\showFig{coltran}{0}{15} \showFig{coltran}{0}{0}}
    \\
    Ground-truth (grayscale)
    & ColTran~\cite{Kumar21Colorization} (colorized)
    & \multicolumn{2}{c}{ColTran~\cite{Kumar21Colorization}}
    & Ground-truth (grayscale)
    & ColTran~\cite{Kumar21Colorization} (colorized)
    & \multicolumn{2}{c}{ColTran~\cite{Kumar21Colorization}} \\[6mm]

    \showFig{ar0}{22}{0}
    & \multicolumn{3}{c}{\showFig{ar9}{22}{0} \showFig{ar9}{22}{1} \showFig{ar9}{22}{17}}
    & \showFig{ar0}{0}{0}
    & \multicolumn{3}{c}{\showFig{ar9}{0}{0} \showFig{ar9}{0}{18} \showFig{ar9}{0}{7}}
    \\
    AR (greedy)
    & \multicolumn{3}{c}{AR (top-p, p=0.9)}
    & AR (greedy)
    & \multicolumn{3}{c}{AR (top-p, p=0.9)} \\[2mm]

    \showFig{nar}{22}{0}
    & \multicolumn{3}{c}{\showFig{cvae}{22}{16} \showFig{cvae}{22}{0} \showFig{cvae}{22}{18}}
    & \showFig{nar}{0}{0}
    & \multicolumn{3}{c}{\showFig{cvae}{0}{8} \showFig{cvae}{0}{0} \showFig{cvae}{0}{11}}
    \\
    NAR
    & \multicolumn{3}{c}{CVAE}
    & NAR
    & \multicolumn{3}{c}{CVAE} \\[2mm]

\end{tabular}}
\end{center}
\caption{Qualitative comparison of color style generation. NAR and CVAE
successfully generate more plausible color styles than the others, and CVAE can
produce multiple variations. Best viewed in color and zoom.}
\label{fig:main_result}
\end{figure*}
\setlength\tabcolsep{6pt}
}

\newcommand{\SuppFigureA}{
\setlength{\tabcolsep}{3pt}
\begin{figure*}
\begin{center}
{\small \begin{tabular}{
    C{20mm}C{20mm}C{20mm}C{20mm}
    C{20mm}C{20mm}C{20mm}C{20mm}
    }
    \showFig{gt}{27}{0}
    & \showFig{mode}{27}{0}
    & \multicolumn{2}{c}{\showFig{sampling}{27}{0} \showFig{sampling}{27}{17}}
    & \showFig{gt}{28}{0}
    & \showFig{mode}{28}{0}
    & \multicolumn{2}{c}{\showFig{sampling}{28}{0} \showFig{sampling}{28}{2}}
    \\
    Ground-truth
    & Stats (mode)
    & \multicolumn{2}{c}{Stats (sampling)}
    & Ground-truth
    & Stats (mode)
    & \multicolumn{2}{c}{Stats (sampling)} \\[2mm]

    \showFig{coltran_gray}{27}{0}
    & \showFig{coltran_color}{27}{0}
    & \multicolumn{2}{c}{\showFig{coltran}{27}{0} \showFig{coltran}{27}{11}}
    & \showFig{coltran_gray}{28}{0}
    & \showFig{coltran_color}{28}{0}
    & \multicolumn{2}{c}{\showFig{coltran}{28}{0} \showFig{coltran}{28}{3}}
    \\
    Ground-truth (grayscale)
    & ColTran~\cite{Kumar21Colorization} (colorized)
    & \multicolumn{2}{c}{ColTran~\cite{Kumar21Colorization}}
    & Ground-truth (grayscale)
    & ColTran~\cite{Kumar21Colorization} (colorized)
    & \multicolumn{2}{c}{ColTran~\cite{Kumar21Colorization}} \\[6mm]

    \showFig{ar0}{27}{0}
    & \multicolumn{3}{c}{\showFig{ar9}{27}{0} \showFig{ar9}{27}{19} \showFig{ar9}{27}{17}}
    & \showFig{ar0}{28}{0}
    & \multicolumn{3}{c}{\showFig{ar9}{28}{0} \showFig{ar9}{28}{7} \showFig{ar9}{28}{11}}
    \\
    AR (greedy)
    & \multicolumn{3}{c}{AR (top-p, p=0.9)}
    & AR (greedy)
    & \multicolumn{3}{c}{AR (top-p, p=0.9)} \\[2mm]

    \showFig{nar}{27}{0}
    & \multicolumn{3}{c}{\showFig{cvae}{27}{0} \showFig{cvae}{27}{4} \showFig{cvae}{27}{13}}
    & \showFig{nar}{28}{0}
    & \multicolumn{3}{c}{\showFig{cvae}{28}{0} \showFig{cvae}{28}{12} \showFig{cvae}{28}{11}}
    \\
    NAR
    & \multicolumn{3}{c}{CVAE}
    & NAR
    & \multicolumn{3}{c}{CVAE} \\[2mm]

\end{tabular}}
\end{center}
\caption{Additional qualitative results (1).}
\label{fig:supp_a}
\end{figure*}
\setlength\tabcolsep{6pt}
}

\newcommand{\SuppFigureB}{
\setlength{\tabcolsep}{3pt}
\begin{figure*}
\begin{center}
{\small \begin{tabular}{
    C{20mm}C{20mm}C{20mm}C{20mm}
    C{20mm}C{20mm}C{20mm}C{20mm}
    }
    \showFig{gt}{26}{0}
    & \showFig{mode}{26}{0}
    & \multicolumn{2}{c}{\showFig{sampling}{26}{0} \showFig{sampling}{26}{17}}
    & \showFig{gt}{20}{0}
    & \showFig{mode}{20}{0}
    & \multicolumn{2}{c}{\showFig{sampling}{20}{0} \showFig{sampling}{20}{2}}
    \\
    Ground-truth
    & Stats (mode)
    & \multicolumn{2}{c}{Stats (sampling)}
    & Ground-truth
    & Stats (mode)
    & \multicolumn{2}{c}{Stats (sampling)} \\[2mm]

    \showFig{coltran_gray}{26}{0}
    & \showFig{coltran_color}{26}{0}
    & \multicolumn{2}{c}{\showFig{coltran}{26}{0} \showFig{coltran}{26}{5}}
    & \showFig{coltran_gray}{20}{0}
    & \showFig{coltran_color}{20}{0}
    & \multicolumn{2}{c}{\showFig{coltran}{20}{0} \showFig{coltran}{20}{19}}
    \\
    Ground-truth (grayscale)
    & ColTran~\cite{Kumar21Colorization} (colorized)
    & \multicolumn{2}{c}{ColTran~\cite{Kumar21Colorization}}
    & Ground-truth (grayscale)
    & ColTran~\cite{Kumar21Colorization} (colorized)
    & \multicolumn{2}{c}{ColTran~\cite{Kumar21Colorization}} \\[6mm]

    \showFig{ar0}{26}{0}
    & \multicolumn{3}{c}{\showFig{ar9}{26}{0} \showFig{ar9}{26}{10} \showFig{ar9}{26}{12}}
    & \showFig{ar0}{20}{0}
    & \multicolumn{3}{c}{\showFig{ar9}{20}{0} \showFig{ar9}{20}{10} \showFig{ar9}{20}{8}}
    \\
    AR (greedy)
    & \multicolumn{3}{c}{AR (top-p, p=0.9)}
    & AR (greedy)
    & \multicolumn{3}{c}{AR (top-p, p=0.9)} \\[2mm]

    \showFig{nar}{26}{0}
    & \multicolumn{3}{c}{\showFig{cvae}{26}{0} \showFig{cvae}{26}{9} \showFig{cvae}{26}{17}}
    & \showFig{nar}{20}{0}
    & \multicolumn{3}{c}{\showFig{cvae}{20}{0} \showFig{cvae}{20}{17} \showFig{cvae}{20}{16}}
    \\
    NAR
    & \multicolumn{3}{c}{CVAE}
    & NAR
    & \multicolumn{3}{c}{CVAE} \\[2mm]

\end{tabular}}
\end{center}
\caption{Additional qualitative results (2).}
\label{fig:supp_b}
\end{figure*}
\setlength\tabcolsep{6pt}
}

\newcommand{\SuppFigureC}{
\setlength{\tabcolsep}{3pt}
\begin{figure*}
\begin{center}
{\small \begin{tabular}{
    C{20mm}C{20mm}C{20mm}C{20mm}
    C{20mm}C{20mm}C{20mm}C{20mm}
    }
    \showFig{gt}{4}{0}
    & \showFig{mode}{4}{0}
    & \multicolumn{2}{c}{\showFig{sampling}{4}{0} \showFig{sampling}{4}{3}}
    & \showFig{gt}{30}{0}
    & \showFig{mode}{30}{0}
    & \multicolumn{2}{c}{\showFig{sampling}{30}{0} \showFig{sampling}{30}{15}}
    \\
    Ground-truth
    & Stats (mode)
    & \multicolumn{2}{c}{Stats (sampling)}
    & Ground-truth
    & Stats (mode)
    & \multicolumn{2}{c}{Stats (sampling)} \\[2mm]

    \showFig{coltran_gray}{4}{0}
    & \showFig{coltran_color}{4}{0}
    & \multicolumn{2}{c}{\showFig{coltran}{4}{0} \showFig{coltran}{4}{13}}
    & \showFig{coltran_gray}{30}{0}
    & \showFig{coltran_color}{30}{0}
    & \multicolumn{2}{c}{\showFig{coltran}{30}{0} \showFig{coltran}{30}{13}}
    \\
    Ground-truth (grayscale)
    & ColTran~\cite{Kumar21Colorization} (colorized)
    & \multicolumn{2}{c}{ColTran~\cite{Kumar21Colorization}}
    & Ground-truth (grayscale)
    & ColTran~\cite{Kumar21Colorization} (colorized)
    & \multicolumn{2}{c}{ColTran~\cite{Kumar21Colorization}} \\[6mm]

    \showFig{ar0}{4}{0}
    & \multicolumn{3}{c}{\showFig{ar9}{4}{0} \showFig{ar9}{4}{17} \showFig{ar9}{4}{15}}
    & \showFig{ar0}{30}{0}
    & \multicolumn{3}{c}{\showFig{ar9}{30}{0} \showFig{ar9}{30}{13} \showFig{ar9}{30}{5}}
    \\
    AR (greedy)
    & \multicolumn{3}{c}{AR (top-p, p=0.9)}
    & AR (greedy)
    & \multicolumn{3}{c}{AR (top-p, p=0.9)} \\[2mm]

    \showFig{nar}{4}{0}
    & \multicolumn{3}{c}{\showFig{cvae}{4}{0} \showFig{cvae}{4}{19} \showFig{cvae}{4}{17}}
    & \showFig{nar}{30}{0}
    & \multicolumn{3}{c}{\showFig{cvae}{30}{0} \showFig{cvae}{30}{7} \showFig{cvae}{30}{2}}
    \\
    NAR
    & \multicolumn{3}{c}{CVAE}
    & NAR
    & \multicolumn{3}{c}{CVAE} \\[2mm]

\end{tabular}}
\end{center}
\caption{Additional qualitative results (3).}
\label{fig:supp_c}
\end{figure*}
\setlength\tabcolsep{6pt}
}

\newcommand{\SuppFigureD}{
\setlength{\tabcolsep}{3pt}
\begin{figure*}
\begin{center}
{\small \begin{tabular}{
    C{20mm}C{20mm}C{20mm}C{20mm}
    C{20mm}C{20mm}C{20mm}C{20mm}
    }
    \showFig{gt}{8}{0}
    & \showFig{mode}{8}{0}
    & \multicolumn{2}{c}{\showFig{sampling}{8}{0} \showFig{sampling}{8}{8}}
    & \showFig{gt}{17}{0}
    & \showFig{mode}{17}{0}
    & \multicolumn{2}{c}{\showFig{sampling}{17}{0} \showFig{sampling}{17}{12}}
    \\
    Ground-truth
    & Stats (mode)
    & \multicolumn{2}{c}{Stats (sampling)}
    & Ground-truth
    & Stats (mode)
    & \multicolumn{2}{c}{Stats (sampling)} \\[2mm]

    \showFig{coltran_gray}{8}{0}
    & \showFig{coltran_color}{8}{0}
    & \multicolumn{2}{c}{\showFig{coltran}{8}{0} \showFig{coltran}{8}{6}}
    & \showFig{coltran_gray}{17}{0}
    & \showFig{coltran_color}{17}{0}
    & \multicolumn{2}{c}{\showFig{coltran}{17}{0} \showFig{coltran}{17}{11}}
    \\
    Ground-truth (grayscale)
    & ColTran~\cite{Kumar21Colorization} (colorized)
    & \multicolumn{2}{c}{ColTran~\cite{Kumar21Colorization}}
    & Ground-truth (grayscale)
    & ColTran~\cite{Kumar21Colorization} (colorized)
    & \multicolumn{2}{c}{ColTran~\cite{Kumar21Colorization}} \\[6mm]

    \showFig{ar0}{8}{0}
    & \multicolumn{3}{c}{\showFig{ar9}{8}{0} \showFig{ar9}{8}{4} \showFig{ar9}{8}{6}}
    & \showFig{ar0}{17}{0}
    & \multicolumn{3}{c}{\showFig{ar9}{17}{0} \showFig{ar9}{17}{14} \showFig{ar9}{17}{8}}
    \\
    AR (greedy)
    & \multicolumn{3}{c}{AR (top-p, p=0.9)}
    & AR (greedy)
    & \multicolumn{3}{c}{AR (top-p, p=0.9)} \\[2mm]

    \showFig{nar}{8}{0}
    & \multicolumn{3}{c}{\showFig{cvae}{8}{0} \showFig{cvae}{8}{12} \showFig{cvae}{8}{5}}
    & \showFig{nar}{17}{0}
    & \multicolumn{3}{c}{\showFig{cvae}{17}{0} \showFig{cvae}{17}{3} \showFig{cvae}{17}{17}}
    \\
    NAR
    & \multicolumn{3}{c}{CVAE}
    & NAR
    & \multicolumn{3}{c}{CVAE} \\[2mm]

\end{tabular}}
\end{center}
\caption{Additional qualitative results (4).}
\label{fig:supp_d}
\end{figure*}
\setlength\tabcolsep{6pt}
}

\newcommand{\SuppFigureE}{
\setlength{\tabcolsep}{3pt}
\begin{figure*}
\begin{center}
{\small \begin{tabular}{
    C{20mm}C{20mm}C{20mm}C{20mm}
    C{20mm}C{20mm}C{20mm}C{20mm}
    }
    \showFig{gt}{32}{0}
    & \showFig{mode}{32}{0}
    & \multicolumn{2}{c}{\showFig{sampling}{32}{0} \showFig{sampling}{32}{16}}
    & \showFig{gt}{5}{0}
    & \showFig{mode}{5}{0}
    & \multicolumn{2}{c}{\showFig{sampling}{5}{0} \showFig{sampling}{5}{1}}
    \\
    Ground-truth
    & Stats (mode)
    & \multicolumn{2}{c}{Stats (sampling)}
    & Ground-truth
    & Stats (mode)
    & \multicolumn{2}{c}{Stats (sampling)} \\[2mm]

    \showFig{coltran_gray}{32}{0}
    & \showFig{coltran_color}{32}{0}
    & \multicolumn{2}{c}{\showFig{coltran}{32}{0} \showFig{coltran}{32}{19}}
    & \showFig{coltran_gray}{5}{0}
    & \showFig{coltran_color}{5}{0}
    & \multicolumn{2}{c}{\showFig{coltran}{5}{0} \showFig{coltran}{5}{17}}
    \\
    Ground-truth (grayscale)
    & ColTran~\cite{Kumar21Colorization} (colorized)
    & \multicolumn{2}{c}{ColTran~\cite{Kumar21Colorization}}
    & Ground-truth (grayscale)
    & ColTran~\cite{Kumar21Colorization} (colorized)
    & \multicolumn{2}{c}{ColTran~\cite{Kumar21Colorization}} \\[6mm]

    \showFig{ar0}{32}{0}
    & \multicolumn{3}{c}{\showFig{ar9}{32}{0} \showFig{ar9}{32}{19} \showFig{ar9}{32}{3}}
    & \showFig{ar0}{5}{0}
    & \multicolumn{3}{c}{\showFig{ar9}{5}{0} \showFig{ar9}{5}{18} \showFig{ar9}{5}{4}}
    \\
    AR (greedy)
    & \multicolumn{3}{c}{AR (top-p, p=0.9)}
    & AR (greedy)
    & \multicolumn{3}{c}{AR (top-p, p=0.9)} \\[2mm]

    \showFig{nar}{32}{0}
    & \multicolumn{3}{c}{\showFig{cvae}{32}{0} \showFig{cvae}{32}{1} \showFig{cvae}{32}{16}}
    & \showFig{nar}{5}{0}
    & \multicolumn{3}{c}{\showFig{cvae}{5}{0} \showFig{cvae}{5}{14} \showFig{cvae}{5}{6}}
    \\
    NAR
    & \multicolumn{3}{c}{CVAE}
    & NAR
    & \multicolumn{3}{c}{CVAE} \\[2mm]

\end{tabular}}
\end{center}
\caption{Additional qualitative results (5).}
\label{fig:supp_e}
\end{figure*}
\setlength\tabcolsep{6pt}
}

\section{Introduction}
Color plays an important role in the visual communication of web pages.  It is
known that colors are associated with certain emotions~\cite{Kaya04Relationship}
and have a significant impact on the trust and satisfaction of
websites~\cite{Cyr10Colour}.  Effectively coloring web pages to achieve the
design goals is a difficult task, as it requires understanding the theories and
heuristics about colors and their combinations~\cite{GapsyStudio20How}.  Another
reason for the difficulty lies in the implicit and practical requirements within
a page, such as overall balance, contrast, and differentiation of color
connotations.  Our aim is to overcome these difficulties with data-driven
modeling to facilitate new applications such as efficient design exploration for
designers, automated design for non-designers, and automated creation of landing
pages for advertised products.

Despite its great industrial potential, the lack of established benchmarks and
the need for extensive domain knowledge for data collection make it difficult to
apply data-driven methodologies to web page coloring, which may explain why
there are fewer related studies in the literature.  Existing methods for
coloring web pages either generate color styles for some specific
elements~\cite{Zhao18Modeling, Qiu22Intelligent} or require already designed web
pages~\cite{Volkova16Automatic, Gu16Data}, and no method has been investigated
that can add plausible color styles to all elements from scratch.  Also, the
datasets used in these studies are not publicly available, making follow-up
studies difficult.  The image colorization
techniques~\cite{Anwar22Image,Kumar21Colorization} can be applied to web page,
however, they require an input image and the extraction of structural color
information from the output image, which narrows the applicable scenarios and
would cause extraction errors that degrade quality.

In this work, we construct a new dataset to take the study of web page coloring
one step further.  The challenges in data collection are that raw web pages have
many redundant elements that do not affect their appearance, and the color
styles defined in their styling sheets are ambiguous as the same color can be
represented differently.  To address these challenges, we use a common web
browser to simplify web pages and retrieve the browser's resolved styles as
canonical color styles.

The coloring problem is then formalized as a task of generating plausible color
styles for web pages provided with content information and hierarchical
structure of the elements.  We present several Transformer-based methods adapted
to this task.  The hierarchical relationship between elements affects the color
style, since child elements are basically rendered on top of their parent
elements.  Our hierarchical modeling to capture these characteristics is
achieved by contextualizing the features with message
passing~\cite{Gilmer17Neural} before feeding them into the Transformer network.

To evaluate the generated color styles in terms of quality and diversity, we
adapt the metric used in image generation tasks~\cite{Heusel17GANs} to ours and
test whether the resulting pages satisfy the standard contrast criterion.
Quantitative and qualitative results demonstrate the advantages of our methods
over statistical and image colorization methods~\cite{Kumar21Colorization}.

Our contributions are summarized as follows:
\begin{itemize}
    \item A dataset of e-commerce mobile web pages tailored for automatic
    coloring of web pages.
    \item A problem formulation of the web page coloring and
    Transformer~\cite{Vaswani17Attention}-based methods that capture
    hierarchical relationships between elements through structural message passing.
    \item Extensive evaluation including comparison with a state-of-the-art image
    colorization method~\cite{Kumar21Colorization}.
\end{itemize}

\section{Related Work}

\subsection{Image Recoloring and Colorization}
The problem of automatically determining color has been extensively addressed,
especially for images.  There are two common tasks: \emph{recoloring}, which
applies different colors to a color image, and \emph{colorization}, which
applies colors to a grayscale image.  For both tasks, the reference-based
approach has been investigated, which is achieved by finding a mapping between
the color or luminance of the source and the color of a given reference image or
palette~\cite{Chang15Palettebased, Peyre19Computational, Bohra20ColorArt,
Kim21Dynamic}.  Volkova~\etal~\cite{Volkova16Automatic} adapted this approach
for recoloring a web page with a reference image.

The reference-based approach requires additional user input, whereas the
approach that uses color knowledge learned from data does not.  Earlier work in
this direction used relatively simple models to learn cost functions, such as
color compatibility, and then colorize images or 2D patterns through
optimization~\cite{ODonovan11Color,Lin13Probabilistic}.  Recently, deep
learning-based methods have been actively studied, especially for the image
colorization task~\cite{Anwar22Image}, and ColTran~\cite{Kumar21Colorization}
that employs the Transformer network~\cite{Vaswani17Attention} as its base model
has demonstrated state-of-the-art performance.

There are two concerns in applying image colorization techniques to web pages.
The first one is that the input is an image, which narrows the applicable
scenarios, and the second is that the quality could be degraded by errors in
extracting the colors from the output image to be used for the web page.  In our
problem formulation, these concerns do not arise because we treat web page
coloring as an estimation problem on structural data.

\subsection{Data-driven Applications for Web Pages}
Research on data-driven applications for web pages has been actively studied in
recent years in different communities, including question
answering~\cite{Chen21WebSRC}, semantic labeling of
elements~\cite{Hotti21Klarna}, information retrieval~\cite{Overwijk22ClueWeb22},
code generation from a screenshot~\cite{Beltramelli18Pix2code}, layout
generation~\cite{Kikuchi21Modeling}, and GUI photomontage~\cite{Zhao21GUIGAN}.
An earlier work on data-driven web page styling is 
Webzeitgeist~\cite{Kumar13Webzeitgeist}, a platform that provides several data
and demographics for over 100,000 web pages, but is no longer publicly
available.

Relatively few studies have been done for the automatic coloring.
Gu~\etal~\cite{Gu16Data} propose an optimization-based coloring framework by
combining several models, including a color contrast model learned using 500 web
pages, and confirm its effectiveness in two tasks: coloring with a prepared
palette and fine-tuning a color design.  Zhao~\etal~\cite{Zhao18Modeling}
address the task of estimating font attributes of an element, including font
color, from the surrounding context.  Most recently,
Qiu~\etal~\cite{Qiu22Intelligent} propose a Transformer-based color
recommendation model for landing pages that is trained with the masked
prediction and the advertisement performance prediction. Their model uses
multiple color palettes extracted from specific regions of annotated screenshots
and can recommend other colors for the target region.

In summary, there is no established method for adding plausible color styles to
all elements from scratch.  There is also no dataset available to study such a
method, therefore, we begin with the construction of a dataset on the structural
coloring of web pages.

\newcommand{\showSample}[1]{\frame{\includegraphics[width=20mm]{samples/#1}}}

\setlength{\tabcolsep}{1pt}
\renewcommand{\arraystretch}{.6}
\begin{figure}
    \begin{center}
        \begin{tabular}{cccc}
            \showSample{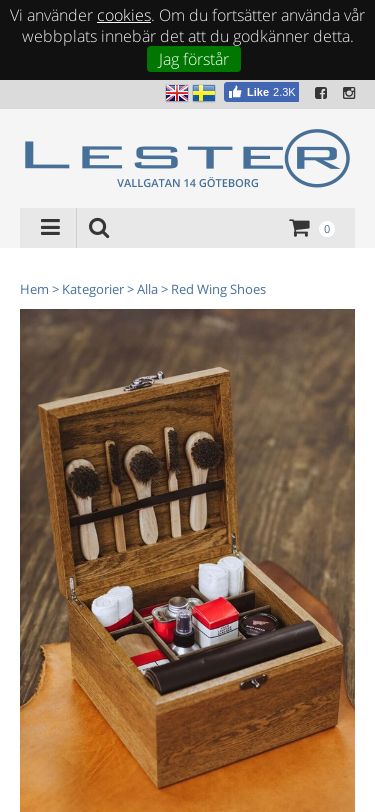} &
            \showSample{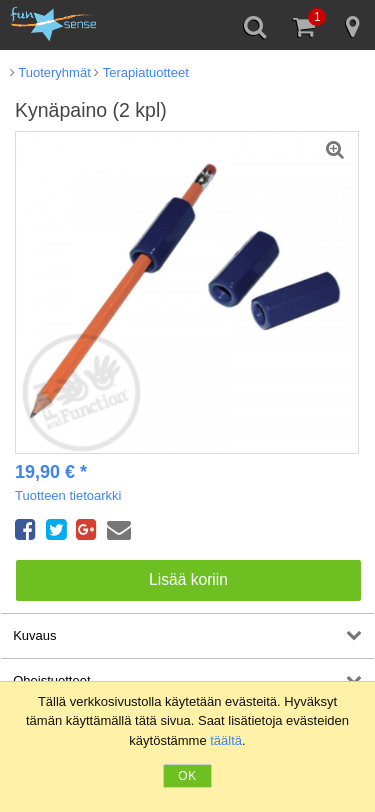} &
            \showSample{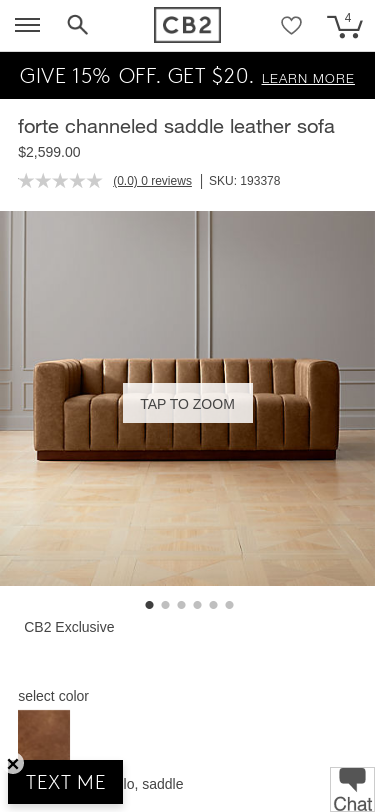} &
            \showSample{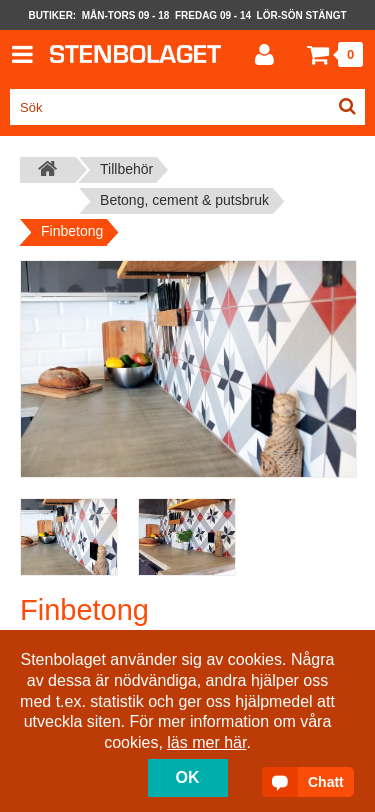}
        \end{tabular}
        \caption{Screenshots of randomly selected web pages.}
        \label{fig:random_samples}
    \end{center}
\end{figure}
\renewcommand{\arraystretch}{1}
\setlength\tabcolsep{6pt}

\begin{figure}
    \begin{center}
        \includegraphics[width=.95\linewidth]{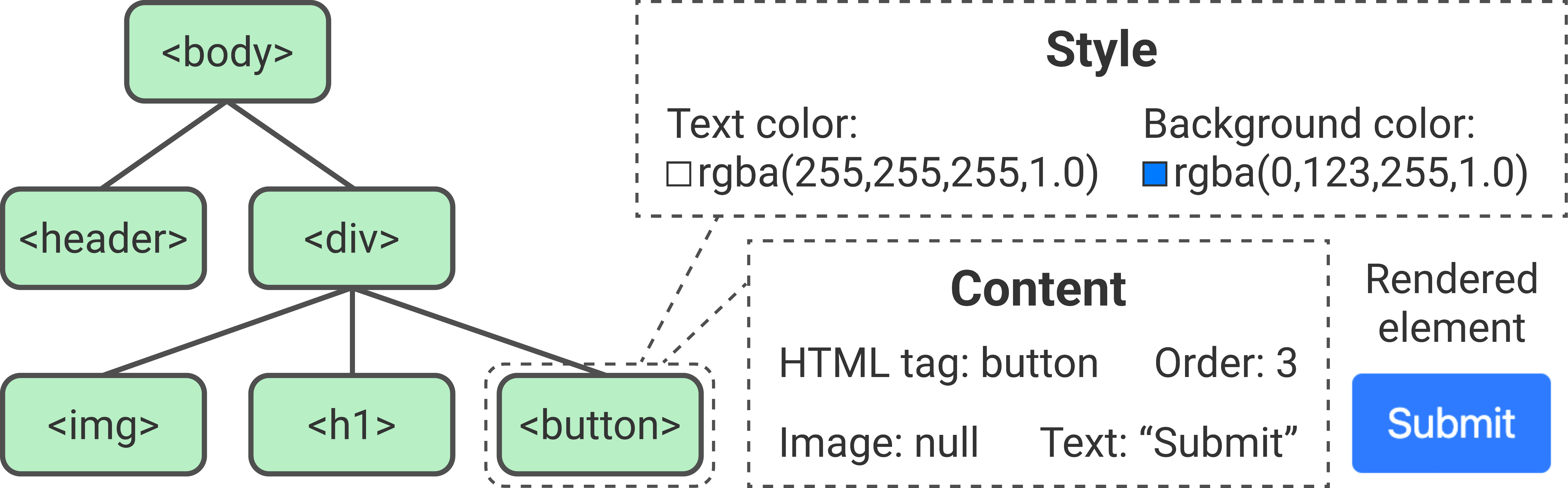}
        \caption{Data format of a web page in this study.}
        \label{fig:data_format}
    \end{center}
\end{figure}

\section{Dataset Construction for Web Page Coloring}
We first describe a generic preprocessing technique that converts web pages into
a tractable data format for machine learning applications.  We then describe the
details of the dataset we have constructed.  Screenshots of randomly selected
web pages from our dataset are shown in \cref{fig:random_samples}.

\subsection{Data Format and Preprocessing}
We represent a web page as an ordered tree, where each vertex is an element to
be rendered on the page and has content and color style information
(\cref{fig:data_format}).  We use HTML tags of elements, their order in siblings
in the tree, and low-level features of images (\eg, average color) and text
(\eg, word counts) as content information, please refer to the supplemental
material for details.  For color styles, it is not trivial to define what to
extract.  Also, raw web pages contain many redundant elements for training the
coloring model.  We discuss below our color style definition and simplification
technique.

\subsubsection{Color Style Definition}
Style information for a web page is defined in Cascading Style Sheets (CSS),
which basically consists of the specifications of target elements (\ie,
selector) and styles to be applied (\ie, properties and their values).  Here we
consider two representative properties \textit{color} and
\textit{background-color} as the color style, corresponding to the text and
background colors of the element, respectively.  There are two possible choices
for where to obtain the values of these properties: specified or computed
values.

The specified values are the values defined in the style sheets.  For elements
where values are not specified, browser's default CSS is applied.  Available
formats for the values of color properties include RGBA values, color keywords,
and special keywords, such as ``inherit'', which causes the use of the same
value as the parent element. Thus, the same color in appearance can be
represented in different formats, and canonicalizing them requires extra effort.
The computed values, on the other hand, are the values resolved by browser.  For
our target properties, the values are disambiguated to RGBA format, which can be
used as the specified values without any change in appearance.  We use the
computed values as the canonical values for the target properties, and we
additionally use the ground-truth specified values for the other properties for
visualization.

\subsubsection{Web Page Simplification}
Raw web pages contain many elements that do not contribute to their appearance
on the present screen, including alternative elements such as elements that only
appear on laptop-sized screens and functional elements such as hidden input
elements in submission forms.  Many redundant or less relevant elements should
be eliminated, as they will negatively affect the learning of the coloring
model.

Here we consider keeping only those elements that contribute to the first view
on the mobile screen.  A naive simplification method would be to try removing an
element and if the appearance of the first view does not change, then really
remove that element.  Removing elements, however, often breaks the original
style when other elements are involved in the CSS selector such as ``div
\textgreater{} a''.  To avoid such undesirable style corruption, before
simplification, we change the CSS to a different one that uses absolute
selectors such as ``\#element-1'' and is carefully tailored to be equivalent to
the original style specification.  As a result, the average number of elements,
excluding those placed in the head element, was significantly reduced from
1656.1 to 61.4.  Note that color styles applied to a simplified web page can be
applied to the original one if the correspondence of the elements is recorded
before and after the simplification.

\subsection{Our Dataset}
The dataset should have a large enough and diverse set of web pages.  We select
the Klarna Product Page dataset~\cite{Hotti21Klarna} as the source of web pages,
which provides mobile web pages for e-commerce in various countries and
companies.  Specifically, we use snapshots of the web pages provided in MHTML
format, which can be replayed in a browser almost identically to the original
pages.  Although the procedures described in this section apply to the
snapshots, they are also applicable to web pages crawled in a general way.

We implement the process of the retrieval of the computed values and the
simplification of web pages using Selenium WebDriver~\cite{SeleniumWebDriver}
and Google Chrome~\cite{GoogleGoogle}.  Complex web pages that, even after
simplification, still have more than 200 elements or a tree depth of more than
30 are excluded.  We also exclude web pages encoded in anything other than utf-8
to simplify text processing.  As a result, a total of 44,048 web pages are left
with an average number of elements of 60.6 and an average tree depth of 9.3.
The web pages of the official training split is divided 9:1 for training and
validation, and those of the official test split for testing, resulting in
27,630 pages for training, 3,190 pages for validation, and 13,228 pages for
testing.

\begin{figure*}[t]
    \begin{center}
        \includegraphics[width=.99\linewidth]{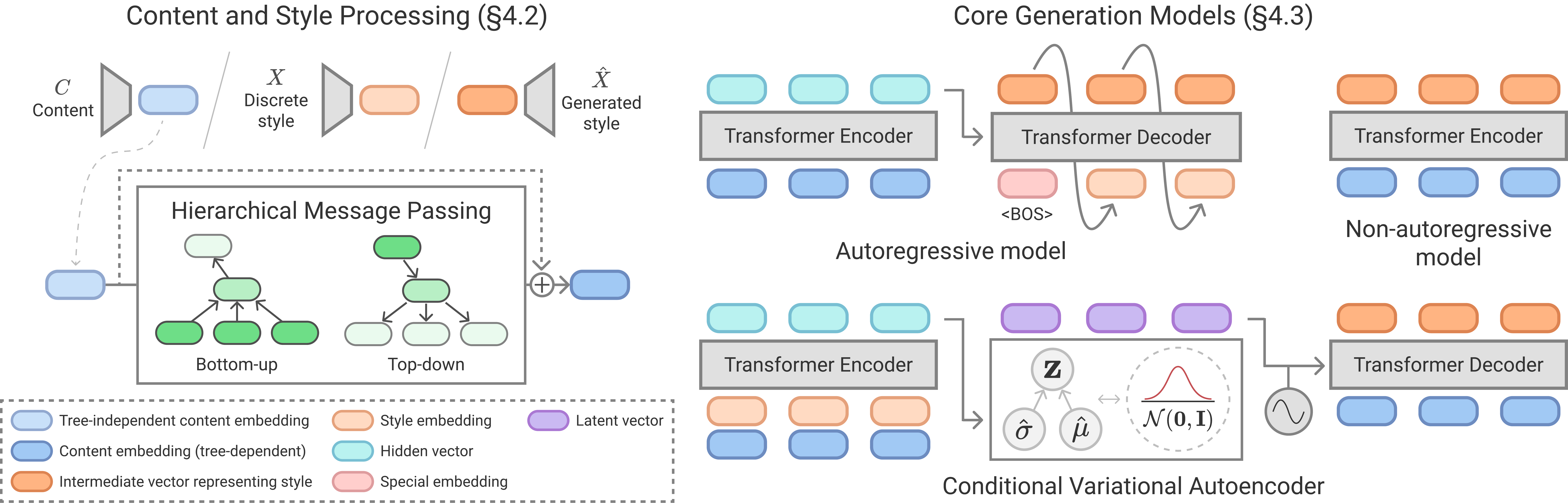}
        \vspace*{-1mm}
        \caption{Key components of our method for generating discrete color
        styles.  Our method has three variants with different core generation
        models, all using the same architecture of the content and style
        encoders and the style estimation head.  Element-wise content features
        are contextualized by features from other elements through hierarchical
        message passing.}
        \label{fig:overview}
    \end{center}
\end{figure*}

\section{Approach} \label{sec:our_methods}
We provide high-level explanations of our approach for the sake of clarity, and
details on message passing, model architecture, and implementation are presented
in the supplemental material.

\subsection{Overview}
We define web page colorization as the task of generating a color style for each
element, given the content information and the hierarchical structure of the
elements.  We denote the index of an element as $n$, the number of elements on
the page as $N$, the set of color style as $\mathcal{Y}=\{Y_n\}_{n=1}^N$, the
set of content information as $\mathcal{C}=\{C_n\}_{n=1}^N$, and the
hierarchical structure as $T$, respectively.  Thus, the main objective of this
task can be formulated as creating a function~$f$ that generates color styles
based on given conditions, \ie, $f: \left( \mathcal{C}, T \right) \mapsto
\hat{\mathcal{Y}}$, where a variable with the hat symbol, \ie
$\hat{\mathcal{Y}}$, represents the estimate of that variable, \ie,
$\mathcal{Y}$.

Following the recent image colorization method~\cite{Kumar21Colorization}, we
take a two-stage approach using a core model to generate low-resolution discrete
styles $\mathcal{X}=\{X_n\}_{n=1}^N$, \ie, $g: \left( \mathcal{C}, T \right)
\mapsto \hat{\mathcal{X}}$, and a color upsampler to increase the resolution of
the colors to the desired size, \ie, $h: \left( \mathcal{X}, \mathcal{C}, T
\right) \mapsto \hat{\mathcal{Y}}$.  For quantization, we divide the RGBA values
into eight bins per channel, and the RGB channels are grouped together.  The
discrete text color of an element is then represented by a pair of 
$\left(x_\mathrm{rgb} \in \{ 1,\dots,8^3 \}, x_\mathrm{alpha} \in \{ 1,\dots,8
\} \right)$.  The same quantization is applied to the background color.  We
consider three core generation models: autoregressive model, non-autoregressive
model, and conditional variational autoencoder, which are all based on
Transformer networks and have the same architecture for the content and style
encoders and the style estimation head.  The core model and the color upsampler
are trained independently.  Some key components of our method are shown in
\cref{fig:overview}.

\subsection{Content and Style Processing} \label{sec:encoders}
The content encoder takes content information for each element
$\{C_{n}\}^N_{n=1}$ (the index $n$ is omitted hereafter) as input, and referring
to the tree structure $T$, converts them into $d$-dimensional content embeddings
$\{\mathbf{h}_C\}$ that reflect the hierarchical relationship of the elements. 
We implement the content encoder using bottom-up and top-down message passing
with residual connections, which is expressed as:
\begin{align}
    \bar{\mathbf{h}}_C &= \mathrm{MaxPool}\left( \{ \mathrm{Emb}_c \left( c
    \right) | c \in C \} \right), \\
    \{ \mathbf{h}_\mathrm{up} \} &= \mathrm{MP}_\mathrm{up}( \{
    \bar{\mathbf{h}}_C \}, \mathbf{h}_\mathrm{leaf}; T), \\
    \{ \mathbf{h}_\mathrm{down} \} &= \mathrm{MP}_\mathrm{down}( 
    \{\mathbf{h}_\mathrm{up}\}, \mathbf{h}_\mathrm{root}; T), \\
    \mathbf{h}_C &= \bar{\mathbf{h}}_C + \mathbf{h}_\mathrm{down},
\end{align}
where $\mathrm{MaxPool}(\cdot)$ is the max-pooling operator,
$\mathrm{Emb}_c(\cdot)$ is the embedding function corresponding to the content
$c$, $\mathbf{h}_\mathrm{leaf}$ and $\mathbf{h}_\mathrm{root}$ are learnable
parameters, respectively.
In the bottom-up message passing $\mathrm{MP}_\mathrm{up}$, the value
$\mathbf{h}_\mathrm{up}$ of an element is computed using the value
$\bar{\mathbf{h}}_\mathrm{C}$ of the element and the values
$\mathbf{h}_\mathrm{up}$ of the elements children or $\mathbf{h}_\mathrm{leaf}$
if the element is a leaf node.  The top-down message passing
$\mathrm{MP}_\mathrm{down}$ is defined similarly in the reverse direction.

The style encoder maps a discrete color style $X$ to a style embedding
$\mathbf{h}_X \in \mathbb{R}^d$ independently for each element.  Specifically,
we represent an RGBA color by a vector obtained by merging two embeddings
corresponding to discrete RGB and alpha values.  The style embedding is obtained
by merging two color vectors corresponding to text and background colors.  For
elements without text, we use a special learnable embedding instead of the text
color vector.  The style estimation head is the module in the final output part
that maps intermediate vectors representing element-by-element styles to
discrete RGB and alpha probabilities for text and background colors,
respectively.

\subsection{Core Generation Models}
The role of the core generation model is to generate a discrete color style for
each element $\{\hat{X}\}$ based on the given conditions.  Here, we implement
this model using the Transformer network and train it with two different
schemes: maximum likelihood estimation (MLE) and conditional variational
autoencoder (CVAE).  The model learned with the MLE is further subdivided into
two variants, autoregressive and non-autoregressive models, resulting in a total
of three variants of the core model to be considered.

\subsubsection{MLE-based Model}
The models are trained by maximizing the log-likelihood.  Let $\theta$ be the
model parameters, the objective is written as
\begin{equation}
    \max_{\theta} E
    \left[ \log p_\theta \left( \mathcal{X} \,|\, \mathcal{C}, T \right) \right].
\end{equation}

\paragraph{Autoregressive Model:}
In the autoregressive model, the conditional probability of the color styles is
modeled as
\begin{equation}
    p_\theta \left( \mathcal{X} \,|\, \mathcal{C}, T \right) \defeq
    \prod_{n=1}^{N} p \left( X_n \,|\, X_1, X_2, \cdots, X_{n-1}, \mathcal{C}, T \right).
\end{equation}
The order of the elements is defined by the pre-order tree traversal.

\paragraph{Non-autoregressive Model:}
In the non-autoregressive model, the conditional probability is assumed to be
not conditioned on the previous estimates and is modeled as
\begin{equation}
    p_\theta \left( \mathcal{X} \,|\, \mathcal{C}, T \right) \defeq
    \prod_{n=1}^N p \left( X_n \,|\, \mathcal{C}, T \right).
\end{equation}

We implement these models with the Transformer networks as illustrated in the
upper right of \cref{fig:overview}. The content and style embeddings and the
style estimation head used are those described in \cref{sec:encoders}.

\subsubsection{CVAE-based Model}
In general, the non-autoregressive model is capable of faster inference than the
autoregressive model, but is unable to create a diverse colorizations for fixed
input values.  We introduce latent variables into the non-autoregressive model
and extend it to the formulation of conditional variational autoencoder,
allowing for the generation of diverse outputs from a single input.

Let us denote latent variables as $Z \in \mathbb{R}^{Nd}$, the conditional
generative distribution as $\vlike$, the posterior as $\vpost$, and the prior as
$\vpri$, respectively, the learning objective of CVAE is expressed as:
\begin{align}
    \max_{\phi, \theta} \quad & E_{\vpost} \left[ \log \vlike \right] \nonumber \\
    & - \lambda \mathrm{KL} \left( \vpost \,\|\, \vpri \right), \label{eq:cvae_objective}
\end{align}
where $\phi$ and $\theta$ are the model parameters and $\lambda$ is a
hyper-parameter to balance the two terms. We set $\lambda=0.1$ for all
experiments.

Using multivariate Gaussian distributions with diagonal covariance matrices, we
model the conditional distributions as follows:
\begin{align}
    \vpost \defeq& \mathcal{N} \left( \vmu, I \vsigma \right) \label{eqn:cvae_post}, \\
    \vlike \defeq& \prod_{n=1}^N p \left( X_n \,|\, Z, \mathcal{C}, T \right), \\
    \vpri \defeq& \mathcal{N} \left( \mathbf{0}, I \right),
\end{align}
where $\mu(\cdot)$ and $\sigma(\cdot)$ are the functions that return parameters
corresponding to the mean and the variance of the Gaussian distribution,
respectively.

We implement this model with both the Transformer encoder and decoder as shown
in the bottom right of \cref{fig:overview}.  The Transformer encoder takes
content and style embeddings as input and produces the estimated mean and the
estimated variance for each element. They are concatenated for all elements and
treated as the returned vectors of $\mu(\cdot)$ and $\sigma(\cdot)$ in
\cref{eqn:cvae_post}, respectively.  The latent variables $Z$ are then sampled
from the Gaussian distribution using the reparametrization trick and partitioned
into a set of latent vectors equal to the number of elements.  The Transformer
decoder takes the latent vectors and embeddings as input and estimates the color
styles of all elements.  Note that the latent vectors are applied the positional
encoding so that the decoder can identify which latent vector corresponds to
which element.

\subsection{Color Upsampler} \label{sec:color_upsampler}
The color upsampler takes the discrete color styles $\mathcal{X}$ as input and
generates the color styles in full resolution $\hat{\mathcal{Y}}$.  To force the
upsampled colors to stay in their original quantization bins, we estimate the
proportions in the bins instead of directly estimating the full-resolution
colors.  We train the Transformer-based model to minimize the mean squared error
between the predicted proportions and the ground-truth proportions.

\section{Experiments}
We evaluate our methods against additional baseline methods in terms of the
quality and diversity of the generated color styles.

\subsection{Methods}
We employ baseline methods based on statistics of the dataset and image
colorization.

\paragraph{Statistics-based Coloring:}
We use simpler methods based on statistics of color styles.  Specifically, we
first collect the frequencies of the discrete color for each pair of HTML tags
and CSS properties in the training set.  The color is then determined by mode
selection (\emph{mode}) or frequency-weighted sampling (\emph{sampling}) and
upsampled in the same way as our methods.  When sampling, we set the same color
for the same pair of tags and properties, encouraging consistent color styling
within a single web page.  For pairs that appear in the test set but not in the
training set, the global frequencies across all the tags are referenced.

\paragraph{Image Colorization~\cite{Kumar21Colorization}:}
We adapt the image colorization technique to our task as an additional baseline.
The main consideration of adaptation is that both input and output are images
rather than structural color information.  We use a screenshot of ground-truth
web page converted to a grayscale image as the input.  Obtaining structural
colors from the output colorized image is not trivial, as it requires knowing
the correspondences between the pixel colors and the element color styles.
Inspired by chroma key compositing, we assign an unused color to a target
element, render the web page, and consider the pixels with that color to be the
corresponding pixels.  The colorized image is then referenced and the most
dominant color in the corresponding pixels is taken for the target element.  We
repeat this process for all the elements.  Note that in this process, all colors
are treated as opaque colors, \ie, the alpha value is $1.0$.  For the specific
method, we employ \emph{ColTran}~\cite{Kumar21Colorization}, a state-of-the-art
method based on Transformer, whose official implementation is publicly
available\footnote{\url{https://github.com/google-research/google-research/tree/master/coltran}}.

\paragraph{Our Methods:}
We use three methods with different core models: the autoregressive model (AR),
the non-autoregressive model (NAR), and the conditional variational autoencoder
(CVAE) described in \cref{sec:our_methods}.  Each method uses the same color
upsampler and the same hyperparameters of the Transformer network, the details
can be found in the supplemental material.  To generate with the AR model, we
use the same trained model with three different decoding strategies: greedy
decoding, top-p sampling~\cite{Holtzman20Curious} with p=0.8 and p=0.9.  Note
that during training, we exclude the text color of elements without text content
from the loss calculation because it does not affect the appearance.

\subsection{Evaluation Metrics}

\paragraph{Accuracy and Macro F-score:}
As proxy metrics for the quality of the generated results, the reproducibility
of the ground-truth data is measured.  We compute the accuracy and the Macro
F-score using the discrete RGB and alpha values. The text color of elements
without text content is excluded from the computation. For the Macro F-score,
the average of class-wise F-scores is used.

\paragraph{Fr\'echet Color Distance:}
The diversity of the generated results is another important measure.  We devise
a new metric, named Fr\'echet Color Distance (FCD), with reference to
FID~\cite{Heusel17GANs}, which is widely used in the image generation and
colorization tasks.  In the FID, the distance between the distributions of the
generated data and the real data is computed using intermediate features of the
pre-trained Inception Network.  These high-level image features may not
represent well low-level color information, thus in the FCD, histograms of the
discrete RGB colors are used as the features to compute the distribution
distance.  The histograms are normalized by the number of elements, and their
statistics are calculated for background colors, text colors, and pixels,
respectively, so that the diversity of the different perspectives can be
measured.  Note that the text color of elements without text content is excluded
from the corresponding histogram.

\paragraph{Contrast violation:}
The quality of the generated results is evaluated from another perspective:
accessibility.  To investigate the accessibility of the generated results, we
use Lighthouse~\cite{Google22Lighthouse}, a commonly used auditing tool in
practice, and consider the contrast audit, which is greatly affected by color
styles.  The contrast audit is based on Web Content Accessibility Guidelines
2.1~\cite{Kirkpatrick18Web} and tests whether the contrast ratio between the
background and text colors meets the criteria for all the text on a page.  We
report the percentage of pages and the average number of elements that violate
the contrast audit.

\MainFigure{}
\MainTable{} 

\subsection{Color Style Generation}
We summarize the quantitative results in~\cref{tab:main_result} and the
qualitative results in~\cref{fig:main_result}.  In the quantitative results, we
report the mean and standard deviation of one evaluation for each of the three
models trained with different random seeds, and for Stats (sampling), we report
them for three evaluations.  The qualitative results show screenshots of web
pages where the generated color styles are applied.  For those with multiple
results, we first generate 20 variations with a single model, then the one with
the largest color distance is greedily selected, starting from a random
selection.

The Fr\'echet Color Distances (FCDs) are calculated between two halves of the
test set, one from the generated results and the other from the ground-truth
data.  The FCDs for real data use the other ground-truth data instead of the
generated results, allowing for reasonable comparisons with other methods.  The
real data shows a large percentage of web pages violating the contrast
criterion, but this is roughly consistent with the public
statistics~\cite{TheHTTPArchiveHTTP}.  We consider the metrics of contrast
violation only for reference, and consider them good if they do not differ
significantly from the real statistics.

The image colorization baseline ColTran~\cite{Kumar21Colorization} performs well
for the Pixel-FCD and poorly for the other metrics, which may indicate that by
using the ground-truth grayscale screenshots, ColTran can generate plausible
color styles only for those with many corresponding pixels, such as the
background color of the body element.  In color style extraction from the
colorized image, the extraction errors for those with fewer corresponding
pixels, such as the color for smaller text as in the right example in
\cref{fig:main_result}, may be the cause of contrast violation. Due to not
considering the alpha composition in the extraction, RGB values obtained
differently from the ground-truth may cause poor accuracies, BG-FCD, and
Text-FCD.

The statistical baselines of structural coloring show higher accuracies and
lower macro F-scores, which indicates the imbalance of color styles where
typical color styles, such as black text and white background, appear much more
frequently than other color styles.  The Stats (sampling) improves BG-FCD and
Text-FCD and degrades Pixel-FCD compared to Stats (mode).  The former is as
expected, and the latter can be interpreted because typical color styles are
often assigned to those with many corresponding pixels in the ground-truth data.
For contrast violation, the results indicate that assigning typical color styles
leads to fewer violations, while sampling background and text colors
independently leads to more violations.

Our Transformer-based methods generally perform better than the other baselines,
and the trends in contrast violations are similar to the real data.  In AR,
accuracies and FCDs show that sampling method can control the typicality of the
output.  Both NAR and CVAE outperform the AR variants with comparable scores and
show better visual results than the others.

\paragraph{Ablation study:}
We perform an ablation study to investigate the performance contribution of the
design of the hierarchical contextualization.  The results using CVAE as the
base model are summarized in the lower part of~\cref{tab:main_result} and show
that removing message passing degrades performance, suggesting the importance of
usage of hierarchical modeling.  We can also see that the performance is
degraded without using the residual connection, indicating that they may help
propagate only the minimum information in complex message passing.

\section{Limitations and Discussion}

While our methods succeed in generating relatively plausible color styles, they
are still far from perfect.  A typical failure case is the functional coloring
that informs the product color or the current slide in a slideshow (\ie,
carousel), as can be found in~\cref{fig:main_result}.  The challenge with the
functional coloring stems from the fact that necessary information is removed by
the simplification or is not included on the page in the first place.  The
necessary information is the representative color to distinguish variations for
the product color, and the order of the displayed slide for the slideshow.  To
address this issue, explicit information needs to be added as additional content
or the user needs to modify it in post-processing.

Elements with certain styles are inevitably associated with specific color
styles. For example, an element with a ``round'' style should have a background
color. However, our methods can fail to handle such styling rules.  Adding
color-related styles could improve the situation, but one needs to consider
which and how much to add.  It may be necessary to find a way to handle all the
complex CSS properties in a unified and efficient manner, which may also lead to
the extension of CSS property generation beyond color.

There is still room for improvement regarding the dataset.  The e-commerce
mobile web pages used in our study have low relevance between individual product
images and color styles, making them unsuitable for evaluating the harmony
between images and colors.  Also, it has not been tested whether our methods can
handle more elements, such as full-page view on a mobile-sized screen or on a
laptop-sized screen.  To facilitate these studies, we believe that the
development of stable data collection tool and the construction of large-scale
datasets using such a tool are promising future directions.

{\small
\bibliographystyle{ieee_fullname}
\bibliography{reference}
}

\end{document}

\title{Generative Colorization of Structured Mobile Web Pages \\
--- Supplemental Material ---}

\author{
  Kotaro Kikuchi$^1$
  \and
  Naoto Inoue$^1$
  \and
  Mayu Otani$^1$\\[.5em]
  \hspace{-8em}$^1$CyberAgent
  \and
  Edgar Simo-Serra$^2$\\[.5em]
  \hspace{-2em}$^2$Waseda University
  \and
  Kota Yamaguchi$^1$
}

\maketitle
\thispagestyle{empty}

\newcommand{\showFig}[3]{\frame{\includegraphics[width=20mm]{results/#1/#2/#3.png}}}

\newcommand{\MainFigure}{
\setlength{\tabcolsep}{3pt}
\begin{figure*}
\begin{center}
{\small \begin{tabular}{
    C{20mm}C{20mm}C{20mm}C{20mm}
    C{20mm}C{20mm}C{20mm}C{20mm}
    }
    \showFig{gt}{22}{0}
    & \showFig{mode}{22}{0}
    & \multicolumn{2}{c}{\showFig{sampling}{22}{0} \showFig{sampling}{22}{2}}
    & \showFig{gt}{0}{0}
    & \showFig{mode}{0}{0}
    & \multicolumn{2}{c}{\showFig{sampling}{0}{0} \showFig{sampling}{0}{5}}
    \\
    Ground-truth
    & Stats (mode)
    & \multicolumn{2}{c}{Stats (sampling)}
    & Ground-truth
    & Stats (mode)
    & \multicolumn{2}{c}{Stats (sampling)} \\[2mm]

    \showFig{coltran_gray}{22}{0}
    & \showFig{coltran_color}{22}{0}
    & \multicolumn{2}{c}{\showFig{coltran}{22}{0} \showFig{coltran}{22}{1}}
    & \showFig{coltran_gray}{0}{0}
    & \showFig{coltran_color}{0}{15}
    & \multicolumn{2}{c}{\showFig{coltran}{0}{15} \showFig{coltran}{0}{0}}
    \\
    Ground-truth (grayscale)
    & ColTran~\cite{Kumar21Colorization} (colorized)
    & \multicolumn{2}{c}{ColTran~\cite{Kumar21Colorization}}
    & Ground-truth (grayscale)
    & ColTran~\cite{Kumar21Colorization} (colorized)
    & \multicolumn{2}{c}{ColTran~\cite{Kumar21Colorization}} \\[6mm]

    \showFig{ar0}{22}{0}
    & \multicolumn{3}{c}{\showFig{ar9}{22}{0} \showFig{ar9}{22}{1} \showFig{ar9}{22}{17}}
    & \showFig{ar0}{0}{0}
    & \multicolumn{3}{c}{\showFig{ar9}{0}{0} \showFig{ar9}{0}{18} \showFig{ar9}{0}{7}}
    \\
    AR (greedy)
    & \multicolumn{3}{c}{AR (top-p, p=0.9)}
    & AR (greedy)
    & \multicolumn{3}{c}{AR (top-p, p=0.9)} \\[2mm]

    \showFig{nar}{22}{0}
    & \multicolumn{3}{c}{\showFig{cvae}{22}{16} \showFig{cvae}{22}{0} \showFig{cvae}{22}{18}}
    & \showFig{nar}{0}{0}
    & \multicolumn{3}{c}{\showFig{cvae}{0}{8} \showFig{cvae}{0}{0} \showFig{cvae}{0}{11}}
    \\
    NAR
    & \multicolumn{3}{c}{CVAE}
    & NAR
    & \multicolumn{3}{c}{CVAE} \\[2mm]

\end{tabular}}
\end{center}
\caption{Qualitative comparison of color style generation. NAR and CVAE
successfully generate more plausible color styles than the others, and CVAE can
produce multiple variations. Best viewed in color and zoom.}
\label{fig:main_result}
\end{figure*}
\setlength\tabcolsep{6pt}
}

\newcommand{\SuppFigureA}{
\setlength{\tabcolsep}{3pt}
\begin{figure*}
\begin{center}
{\small \begin{tabular}{
    C{20mm}C{20mm}C{20mm}C{20mm}
    C{20mm}C{20mm}C{20mm}C{20mm}
    }
    \showFig{gt}{27}{0}
    & \showFig{mode}{27}{0}
    & \multicolumn{2}{c}{\showFig{sampling}{27}{0} \showFig{sampling}{27}{17}}
    & \showFig{gt}{28}{0}
    & \showFig{mode}{28}{0}
    & \multicolumn{2}{c}{\showFig{sampling}{28}{0} \showFig{sampling}{28}{2}}
    \\
    Ground-truth
    & Stats (mode)
    & \multicolumn{2}{c}{Stats (sampling)}
    & Ground-truth
    & Stats (mode)
    & \multicolumn{2}{c}{Stats (sampling)} \\[2mm]

    \showFig{coltran_gray}{27}{0}
    & \showFig{coltran_color}{27}{0}
    & \multicolumn{2}{c}{\showFig{coltran}{27}{0} \showFig{coltran}{27}{11}}
    & \showFig{coltran_gray}{28}{0}
    & \showFig{coltran_color}{28}{0}
    & \multicolumn{2}{c}{\showFig{coltran}{28}{0} \showFig{coltran}{28}{3}}
    \\
    Ground-truth (grayscale)
    & ColTran~\cite{Kumar21Colorization} (colorized)
    & \multicolumn{2}{c}{ColTran~\cite{Kumar21Colorization}}
    & Ground-truth (grayscale)
    & ColTran~\cite{Kumar21Colorization} (colorized)
    & \multicolumn{2}{c}{ColTran~\cite{Kumar21Colorization}} \\[6mm]

    \showFig{ar0}{27}{0}
    & \multicolumn{3}{c}{\showFig{ar9}{27}{0} \showFig{ar9}{27}{19} \showFig{ar9}{27}{17}}
    & \showFig{ar0}{28}{0}
    & \multicolumn{3}{c}{\showFig{ar9}{28}{0} \showFig{ar9}{28}{7} \showFig{ar9}{28}{11}}
    \\
    AR (greedy)
    & \multicolumn{3}{c}{AR (top-p, p=0.9)}
    & AR (greedy)
    & \multicolumn{3}{c}{AR (top-p, p=0.9)} \\[2mm]

    \showFig{nar}{27}{0}
    & \multicolumn{3}{c}{\showFig{cvae}{27}{0} \showFig{cvae}{27}{4} \showFig{cvae}{27}{13}}
    & \showFig{nar}{28}{0}
    & \multicolumn{3}{c}{\showFig{cvae}{28}{0} \showFig{cvae}{28}{12} \showFig{cvae}{28}{11}}
    \\
    NAR
    & \multicolumn{3}{c}{CVAE}
    & NAR
    & \multicolumn{3}{c}{CVAE} \\[2mm]

\end{tabular}}
\end{center}
\caption{Additional qualitative results (1).}
\label{fig:supp_a}
\end{figure*}
\setlength\tabcolsep{6pt}
}

\newcommand{\SuppFigureB}{
\setlength{\tabcolsep}{3pt}
\begin{figure*}
\begin{center}
{\small \begin{tabular}{
    C{20mm}C{20mm}C{20mm}C{20mm}
    C{20mm}C{20mm}C{20mm}C{20mm}
    }
    \showFig{gt}{26}{0}
    & \showFig{mode}{26}{0}
    & \multicolumn{2}{c}{\showFig{sampling}{26}{0} \showFig{sampling}{26}{17}}
    & \showFig{gt}{20}{0}
    & \showFig{mode}{20}{0}
    & \multicolumn{2}{c}{\showFig{sampling}{20}{0} \showFig{sampling}{20}{2}}
    \\
    Ground-truth
    & Stats (mode)
    & \multicolumn{2}{c}{Stats (sampling)}
    & Ground-truth
    & Stats (mode)
    & \multicolumn{2}{c}{Stats (sampling)} \\[2mm]

    \showFig{coltran_gray}{26}{0}
    & \showFig{coltran_color}{26}{0}
    & \multicolumn{2}{c}{\showFig{coltran}{26}{0} \showFig{coltran}{26}{5}}
    & \showFig{coltran_gray}{20}{0}
    & \showFig{coltran_color}{20}{0}
    & \multicolumn{2}{c}{\showFig{coltran}{20}{0} \showFig{coltran}{20}{19}}
    \\
    Ground-truth (grayscale)
    & ColTran~\cite{Kumar21Colorization} (colorized)
    & \multicolumn{2}{c}{ColTran~\cite{Kumar21Colorization}}
    & Ground-truth (grayscale)
    & ColTran~\cite{Kumar21Colorization} (colorized)
    & \multicolumn{2}{c}{ColTran~\cite{Kumar21Colorization}} \\[6mm]

    \showFig{ar0}{26}{0}
    & \multicolumn{3}{c}{\showFig{ar9}{26}{0} \showFig{ar9}{26}{10} \showFig{ar9}{26}{12}}
    & \showFig{ar0}{20}{0}
    & \multicolumn{3}{c}{\showFig{ar9}{20}{0} \showFig{ar9}{20}{10} \showFig{ar9}{20}{8}}
    \\
    AR (greedy)
    & \multicolumn{3}{c}{AR (top-p, p=0.9)}
    & AR (greedy)
    & \multicolumn{3}{c}{AR (top-p, p=0.9)} \\[2mm]

    \showFig{nar}{26}{0}
    & \multicolumn{3}{c}{\showFig{cvae}{26}{0} \showFig{cvae}{26}{9} \showFig{cvae}{26}{17}}
    & \showFig{nar}{20}{0}
    & \multicolumn{3}{c}{\showFig{cvae}{20}{0} \showFig{cvae}{20}{17} \showFig{cvae}{20}{16}}
    \\
    NAR
    & \multicolumn{3}{c}{CVAE}
    & NAR
    & \multicolumn{3}{c}{CVAE} \\[2mm]

\end{tabular}}
\end{center}
\caption{Additional qualitative results (2).}
\label{fig:supp_b}
\end{figure*}
\setlength\tabcolsep{6pt}
}

\newcommand{\SuppFigureC}{
\setlength{\tabcolsep}{3pt}
\begin{figure*}
\begin{center}
{\small \begin{tabular}{
    C{20mm}C{20mm}C{20mm}C{20mm}
    C{20mm}C{20mm}C{20mm}C{20mm}
    }
    \showFig{gt}{4}{0}
    & \showFig{mode}{4}{0}
    & \multicolumn{2}{c}{\showFig{sampling}{4}{0} \showFig{sampling}{4}{3}}
    & \showFig{gt}{30}{0}
    & \showFig{mode}{30}{0}
    & \multicolumn{2}{c}{\showFig{sampling}{30}{0} \showFig{sampling}{30}{15}}
    \\
    Ground-truth
    & Stats (mode)
    & \multicolumn{2}{c}{Stats (sampling)}
    & Ground-truth
    & Stats (mode)
    & \multicolumn{2}{c}{Stats (sampling)} \\[2mm]

    \showFig{coltran_gray}{4}{0}
    & \showFig{coltran_color}{4}{0}
    & \multicolumn{2}{c}{\showFig{coltran}{4}{0} \showFig{coltran}{4}{13}}
    & \showFig{coltran_gray}{30}{0}
    & \showFig{coltran_color}{30}{0}
    & \multicolumn{2}{c}{\showFig{coltran}{30}{0} \showFig{coltran}{30}{13}}
    \\
    Ground-truth (grayscale)
    & ColTran~\cite{Kumar21Colorization} (colorized)
    & \multicolumn{2}{c}{ColTran~\cite{Kumar21Colorization}}
    & Ground-truth (grayscale)
    & ColTran~\cite{Kumar21Colorization} (colorized)
    & \multicolumn{2}{c}{ColTran~\cite{Kumar21Colorization}} \\[6mm]

    \showFig{ar0}{4}{0}
    & \multicolumn{3}{c}{\showFig{ar9}{4}{0} \showFig{ar9}{4}{17} \showFig{ar9}{4}{15}}
    & \showFig{ar0}{30}{0}
    & \multicolumn{3}{c}{\showFig{ar9}{30}{0} \showFig{ar9}{30}{13} \showFig{ar9}{30}{5}}
    \\
    AR (greedy)
    & \multicolumn{3}{c}{AR (top-p, p=0.9)}
    & AR (greedy)
    & \multicolumn{3}{c}{AR (top-p, p=0.9)} \\[2mm]

    \showFig{nar}{4}{0}
    & \multicolumn{3}{c}{\showFig{cvae}{4}{0} \showFig{cvae}{4}{19} \showFig{cvae}{4}{17}}
    & \showFig{nar}{30}{0}
    & \multicolumn{3}{c}{\showFig{cvae}{30}{0} \showFig{cvae}{30}{7} \showFig{cvae}{30}{2}}
    \\
    NAR
    & \multicolumn{3}{c}{CVAE}
    & NAR
    & \multicolumn{3}{c}{CVAE} \\[2mm]

\end{tabular}}
\end{center}
\caption{Additional qualitative results (3).}
\label{fig:supp_c}
\end{figure*}
\setlength\tabcolsep{6pt}
}

\newcommand{\SuppFigureD}{
\setlength{\tabcolsep}{3pt}
\begin{figure*}
\begin{center}
{\small \begin{tabular}{
    C{20mm}C{20mm}C{20mm}C{20mm}
    C{20mm}C{20mm}C{20mm}C{20mm}
    }
    \showFig{gt}{8}{0}
    & \showFig{mode}{8}{0}
    & \multicolumn{2}{c}{\showFig{sampling}{8}{0} \showFig{sampling}{8}{8}}
    & \showFig{gt}{17}{0}
    & \showFig{mode}{17}{0}
    & \multicolumn{2}{c}{\showFig{sampling}{17}{0} \showFig{sampling}{17}{12}}
    \\
    Ground-truth
    & Stats (mode)
    & \multicolumn{2}{c}{Stats (sampling)}
    & Ground-truth
    & Stats (mode)
    & \multicolumn{2}{c}{Stats (sampling)} \\[2mm]

    \showFig{coltran_gray}{8}{0}
    & \showFig{coltran_color}{8}{0}
    & \multicolumn{2}{c}{\showFig{coltran}{8}{0} \showFig{coltran}{8}{6}}
    & \showFig{coltran_gray}{17}{0}
    & \showFig{coltran_color}{17}{0}
    & \multicolumn{2}{c}{\showFig{coltran}{17}{0} \showFig{coltran}{17}{11}}
    \\
    Ground-truth (grayscale)
    & ColTran~\cite{Kumar21Colorization} (colorized)
    & \multicolumn{2}{c}{ColTran~\cite{Kumar21Colorization}}
    & Ground-truth (grayscale)
    & ColTran~\cite{Kumar21Colorization} (colorized)
    & \multicolumn{2}{c}{ColTran~\cite{Kumar21Colorization}} \\[6mm]

    \showFig{ar0}{8}{0}
    & \multicolumn{3}{c}{\showFig{ar9}{8}{0} \showFig{ar9}{8}{4} \showFig{ar9}{8}{6}}
    & \showFig{ar0}{17}{0}
    & \multicolumn{3}{c}{\showFig{ar9}{17}{0} \showFig{ar9}{17}{14} \showFig{ar9}{17}{8}}
    \\
    AR (greedy)
    & \multicolumn{3}{c}{AR (top-p, p=0.9)}
    & AR (greedy)
    & \multicolumn{3}{c}{AR (top-p, p=0.9)} \\[2mm]

    \showFig{nar}{8}{0}
    & \multicolumn{3}{c}{\showFig{cvae}{8}{0} \showFig{cvae}{8}{12} \showFig{cvae}{8}{5}}
    & \showFig{nar}{17}{0}
    & \multicolumn{3}{c}{\showFig{cvae}{17}{0} \showFig{cvae}{17}{3} \showFig{cvae}{17}{17}}
    \\
    NAR
    & \multicolumn{3}{c}{CVAE}
    & NAR
    & \multicolumn{3}{c}{CVAE} \\[2mm]

\end{tabular}}
\end{center}
\caption{Additional qualitative results (4).}
\label{fig:supp_d}
\end{figure*}
\setlength\tabcolsep{6pt}
}

\newcommand{\SuppFigureE}{
\setlength{\tabcolsep}{3pt}
\begin{figure*}
\begin{center}
{\small \begin{tabular}{
    C{20mm}C{20mm}C{20mm}C{20mm}
    C{20mm}C{20mm}C{20mm}C{20mm}
    }
    \showFig{gt}{32}{0}
    & \showFig{mode}{32}{0}
    & \multicolumn{2}{c}{\showFig{sampling}{32}{0} \showFig{sampling}{32}{16}}
    & \showFig{gt}{5}{0}
    & \showFig{mode}{5}{0}
    & \multicolumn{2}{c}{\showFig{sampling}{5}{0} \showFig{sampling}{5}{1}}
    \\
    Ground-truth
    & Stats (mode)
    & \multicolumn{2}{c}{Stats (sampling)}
    & Ground-truth
    & Stats (mode)
    & \multicolumn{2}{c}{Stats (sampling)} \\[2mm]

    \showFig{coltran_gray}{32}{0}
    & \showFig{coltran_color}{32}{0}
    & \multicolumn{2}{c}{\showFig{coltran}{32}{0} \showFig{coltran}{32}{19}}
    & \showFig{coltran_gray}{5}{0}
    & \showFig{coltran_color}{5}{0}
    & \multicolumn{2}{c}{\showFig{coltran}{5}{0} \showFig{coltran}{5}{17}}
    \\
    Ground-truth (grayscale)
    & ColTran~\cite{Kumar21Colorization} (colorized)
    & \multicolumn{2}{c}{ColTran~\cite{Kumar21Colorization}}
    & Ground-truth (grayscale)
    & ColTran~\cite{Kumar21Colorization} (colorized)
    & \multicolumn{2}{c}{ColTran~\cite{Kumar21Colorization}} \\[6mm]

    \showFig{ar0}{32}{0}
    & \multicolumn{3}{c}{\showFig{ar9}{32}{0} \showFig{ar9}{32}{19} \showFig{ar9}{32}{3}}
    & \showFig{ar0}{5}{0}
    & \multicolumn{3}{c}{\showFig{ar9}{5}{0} \showFig{ar9}{5}{18} \showFig{ar9}{5}{4}}
    \\
    AR (greedy)
    & \multicolumn{3}{c}{AR (top-p, p=0.9)}
    & AR (greedy)
    & \multicolumn{3}{c}{AR (top-p, p=0.9)} \\[2mm]

    \showFig{nar}{32}{0}
    & \multicolumn{3}{c}{\showFig{cvae}{32}{0} \showFig{cvae}{32}{1} \showFig{cvae}{32}{16}}
    & \showFig{nar}{5}{0}
    & \multicolumn{3}{c}{\showFig{cvae}{5}{0} \showFig{cvae}{5}{14} \showFig{cvae}{5}{6}}
    \\
    NAR
    & \multicolumn{3}{c}{CVAE}
    & NAR
    & \multicolumn{3}{c}{CVAE} \\[2mm]

\end{tabular}}
\end{center}
\caption{Additional qualitative results (5).}
\label{fig:supp_e}
\end{figure*}
\setlength\tabcolsep{6pt}
}

\section{Content Information Details}
We represent a web page as an ordered tree, where each vertex is an element on
the page and has content and color style information, as described in
Section~3.1 of the main paper. A list of content information used in our
experiments is shown in \cref{tab:content_detail}.

\renewcommand{\arraystretch}{1.5}
\begin{table}[h] \begin{center}
\caption{
    List of content information and corresponding embedding methods used in our
    experiments.  \emph{Lookup} refers to a lookup translation to learnable
    embeddings, and \emph{dense} refers to a linear mapping.
}
\label{tab:content_detail}

{\small
\begin{tabular}{L{1.9cm} L{8.6cm} L{3.5cm} l} \toprule
Name & Description & Domain & Embedding \\ \midrule

Order &
Element order within their siblings &
$\mathbf{N}$ &
Lookup \\

Tag &
HTML tag &
$\{$ ``html'', ``body'', ``header'', ``button'', $\cdots \}$ &
Lookup \\

Text &
For elements with text, we extract features: number of lines and words,
indicator variables for literal features (uppercase, capitals, numbers, and
etc.), and whether it comes from the pseudo element or not. Zero vector
otherwise. &
$\left( \mathbf{N}^2 + \{0,1\}^9 + \{0,1\} \right)$ or $\{0\}^{12}$ &
Dense \\

Image &
For ``img'' tag elements, we extract features of the image referenced in ``src''
attribute: width, height, channel size, aspect ratio, mean and
standard
deviation in RGBA color, and whether it is SVG or not. Zero vector otherwise. &
$\left( \mathbf{N}^3 + \mathbf{R}^9 + \{0,1\} \right)$ or $\{0\}^{13}$ &
Dense \\

Background image &
Same as above, except the image comes from the computed value of
``background-image'' property. &
$\left( \mathbf{N}^3 + \mathbf{R}^9 + \{0,1\} \right)$ or $\{0\}^{13}$ &
Dense \\

\bottomrule \end{tabular} }
\end{center} \end{table}
\renewcommand{\arraystretch}{1}

\section{Details of Hierarchical Message Passing}
As explained in Section~4.2 and Eq.~(1)-(4) of the main paper, we encode content
information of the elements with the bottom-up and top-down message
passing~\cite{Gilmer17Neural}.  Omitting the element index $n$ for the variables
for sets, he bottom-up message passing, $\mathrm{MP}_\mathrm{up}: \left( \{
\bar{\mathbf{h}}_C \}, \mathbf{h}_\mathrm{leaf}; T \right) \mapsto \{
\mathbf{h}_\mathrm{up} \} $ used in Eq.~(2) of the main paper, is defined as
\begin{align}
    \mathbf{h}_\mathrm{up}^{(n)} &=
    \begin{cases}
        \mathrm{MLP}_\mathrm{up}\left( \bar{\mathbf{h}}_C^{(n)} \oplus
        \mathbf{h}_\mathrm{leaf} \right)
            & \text{if } \mathrm{Child}\left( n; T \right) = \emptyset, \\
        \mathrm{MaxPool}\left( \left\{ \mathrm{MLP}_\mathrm{up}\left(
            \bar{\mathbf{h}}_C^{(n)} \oplus \mathbf{h}_\mathrm{up}^{(c)}
            \right) \right\}_{c \in \mathrm{Child}\left( n; T
        \right)} \right)
            & \text{otherwise},
    \end{cases}
\end{align}
and the top-down message passing,
$\mathrm{MP}_\mathrm{down}: \left( \{ \mathbf{h}_\mathrm{up} \},
\mathbf{h}_\mathrm{root}; T \right) \mapsto \{ \mathbf{h}_\mathrm{down} \} $ used
in Eq.~(3) of the main paper, is defined as
\begin{align}
    \mathbf{h}_\mathrm{down}^{(n)} &=
    \begin{cases}
        \mathrm{MLP}_\mathrm{down}\left( \mathbf{h}_\mathrm{up}^{(n)} \oplus
        \mathbf{h}_\mathrm{root} \right)
            & \text{if } n \text{ is the root element of the tree } T, \\
        \mathrm{MLP}_\mathrm{down}\left( \mathbf{h}_\mathrm{up}^{(n)} \oplus
        \mathbf{h}_\mathrm{down}^{\mathrm{Parent}\left( n; T \right)} \right)
            & \text{otherwise}.
    \end{cases}
\end{align}
In the equations above, we represent a multilayer perceptron as
$\mathrm{MLP}(\cdot)$, the concatenation operator as $\oplus$, and the
operations of extracting the children of a given element from the tree as
$\mathrm{Child}(\cdot)$ and extracting the parent as $\mathrm{Parent}(\cdot)$,
respectively.

\section{Details of Core Generation Models}

\subsection{Autoregressive Model}
We implement the model with both the Transformer encoder and
decoder~\cite{Vaswani17Attention}.  The Transformer encoder takes content
embeddings as input and outputs hidden vectors.  The Transformer decoder takes
as input the embeddings of the estimated styles of the previous elements, and
while attending to the hidden vectors, it estimates the color style for the next
element.  For the first style estimation in the decoder, a special learnable
embedding is used as input instead of a style embedding.  The model is trained
by the teacher forcing~\cite{Williams89Learning}, and at test time, it generates
a color style for an element in one inference and repeats it for all the
elements.

\subsection{Non-autoregressive Model}
We implement the model only with the Transformer encoder, which takes content
embeddings as input and estimates the color styles of all the elements
simultaneously.

\section{Color Upsampler Details}
As explained in Section~4.4 of the main paper, our color upsampler estimates the
proportions in the quantization bins instead of the full resolution colors.  Let
$\boldsymbol{\alpha}$ be a vector representing the ground-truth proportions in
the bins for all colors, the learning objective of the modified color upsampler
model $\tilde{h}: \left( \mathcal{X}, \mathcal{C}, T \right) \mapsto
\hat{\boldsymbol{\alpha}}$ is defined as:
\begin{equation}
    \min_{\psi} E\left[ \left(
        \tilde{h}\left( \mathcal{X}, \mathcal{C}, T ; \psi \right)
        - \boldsymbol{\alpha}
    \right)^2 \right],
\end{equation}
where $\psi$ is the model parameters. We implement this model with the
Transformer encoder, which takes content and style embeddings as input.

\ignore{
$\boldsymbol{\alpha} = [ \alpha_r, \alpha_g, \alpha_b, \alpha_a]^\top \in [0,1]^4$
$\tilde{h} \left( \mathcal{X}, \mathcal{C}, T; \psi \right)$ 
\begin{equation}
    y_\mathrm{r} =
    y_\mathrm{r}^\lfloor\left( x_\mathrm{rgb} \right) +
    \alpha_\mathrm{r} \left(
        y_\mathrm{r}^\lceil\left( x_\mathrm{rgb} \right) -
        y_\mathrm{r}^\lfloor\left( x_\mathrm{rgb} \right) -
        \epsilon
    \right)
\end{equation}
}

\section{Implementation Details}
We set the same hyperparameters for all the Transformer networks: the feature
dimension of input and output is 256, the number of attention heads is 8, the
number of layers is 4, and the dimension of the inner feedforward network is
512.  The layer normalization is performed on each layer before other
operations~\cite{Xiong20Layer}.  We train our models with a batch size of 32 for
100,000 iterations, using the AdamW optimizer~\cite{Loshchilov19Decoupled} with
a learning rate of 1e-4.

\section{Additional Visual Results}
We present additional visual results in
\cref{fig:supp_a,fig:supp_b,fig:supp_c,fig:supp_d,fig:supp_e}.  Overall, the
same trends as in the main paper can be observed: NAR and CVAE produce better
color styles, and CVAE is capable of producing multiple variations.  However,
the results are not yet perfect and have several limitations.  One of the
typical failure cases discussed in Section 6 of the main paper is that elements
with certain styles are not colorized.  The right side example in
\cref{fig:supp_e} shows this failure case, as the icons and text of ``Join as a
Member...'' are defined as rounded elements but no visible background colors are
generated.

\SuppFigureA{}
\SuppFigureB{}
\SuppFigureC{}
\SuppFigureD{}
\SuppFigureE{}

{\small
\bibliographystyle{ieee_fullname}
\bibliography{reference}
}